\documentclass[10pt, a4paper]{article}
\usepackage{lrec2022} % this is the new LREC2022 Style
\usepackage{multibib}
\newcites{languageresource}{Language Resources}
\usepackage{graphicx}
\usepackage{tabularx}
\usepackage{soul}
\usepackage{titlesec}
\titleformat{\section}{\normalfont\large\bfseries\center}{\thesection.}{1em}{}
\titleformat{\subsection}{\normalfont\SmallTitleFont\bfseries\raggedright}{\thesubsection.}{1em}{}
\titleformat{\subsubsection}{\normalfont\normalsize\bfseries\raggedright}{\thesubsubsection.}{1em}{}
\renewcommand\thesection{\arabic{section}}
\renewcommand\thesubsection{\thesection.\arabic{subsection}}
\renewcommand\thesubsubsection{\thesubsection.\arabic{subsubsection}}
\usepackage{epstopdf}
\usepackage[utf8]{inputenc}

\usepackage{hyperref}
\usepackage{xstring}

\usepackage{color}

\usepackage{subcaption}
\usepackage{graphicx}
\usepackage{pgfplots}
\usepackage{pgfplotstable}
\usepackage{subcaption}
\usepackage{amsmath}
\usepackage{tikz}
\usepackage{amsfonts}
\usepackage{siunitx}
\usepackage{booktabs}
\usepackage{caption}
\usepackage{subcaption}
\usetikzlibrary{patterns}
\usepackage{soul}
\usepackage{url}
\usepackage{breqn}
\usepackage{makecell}
\usepackage{tabularx}
\usepackage{bm}
\newcommand{\specialcell}[2][c]{%
\begin{tabular}[#1]{@{}c@{}}#2\end{tabular}}

\title{A Transfer Learning Pipeline for Educational Resource Discovery \\with Application in Leading Paragraph Generation}

\name{Irene Li$^1$, Thomas George$^2$, Alexander Fabbri$^1$, Tammy Liao$^1$, Benjamin Chen$^1$, \\ \textbf{\large{Rina Kawamura$^1$, Richard Zhou$^1$, Vanessa Yan$^1$, Swapnil Hingmire$^3$, Dragomir Radev$^1$}}} 

\address{$^1$Yale University, USA, 
        $^2$University of Waterloo, Canada, 
        $^3$Tata Consultancy Services Limited, India \\
         \{irene.li,alexander.fabbri,tammy.liao,benjamin.chen,rina.kawamura\}@yale.edu\\
         \{richard.zhou,vanessa.yandragomir.radev\}@yale.edu\\
         thomasgeorge1001@gmail.com,         swapnil.hingmire@tcs.com\\}

\abstract{
Effective human learning depends on a wide selection of educational materials that align with the learner's current understanding of the topic. While the Internet has revolutionized human learning or education, a substantial resource accessibility barrier still exists. Namely, the excess of online information can make it challenging to navigate and discover high-quality learning materials. In this paper, we propose the educational resource discovery (ERD) pipeline that automates web resource discovery for novel domains. The pipeline consists of three main steps: data collection, feature extraction, and resource classification.  We start with a known source domain and conduct resource discovery on two unseen target domains via transfer learning. We first collect frequent queries from a set of seed documents and search on the web to obtain candidate resources, such as lecture slides and introductory blog posts. Then we introduce a novel pretrained information retrieval deep neural network model, query-document masked language modeling (QD-MLM), to extract deep features of these candidate resources. We apply a tree-based classifier to decide whether the candidate is a positive learning resource. The pipeline achieves F1 scores of 0.94 and 0.82 when evaluated on two similar but novel target domains. Finally, we demonstrate how this pipeline can benefit an application: leading paragraph generation for surveys. This is the first study that considers various web resources for survey generation, to the best of our knowledge. We also release a corpus of 39,728 manually labeled web resources and 659 queries from NLP, Computer Vision (CV), and Statistics (STATS). 
 \\ \newline  }

\begin{document}

\definecolor{clr_cv}{RGB}{161,116,161}
\definecolor{clr_cv_font}{RGB}{205,174,205}
\definecolor{clr_stats}{RGB}{107,143,179}
\definecolor{clr_stats_font}{RGB}{154,186,218}
\definecolor{clr_nlp}{RGB}{160,160,160}
\definecolor{clr_both_font}{RGB}{224,224,224}

\maketitleabstract

\section{Introduction}

People rely on the internet for various educational activities, such as watching lectures, reading textbooks, articles, and encyclopedia pages. One may wish to develop their knowledge in a familiar subject area or to learn something entirely new. Many online tools enable and promote independent learning \cite{montalvo2018building,romero2017educational,fabbri2018tutorialbank,li2019should}. A subset of these platforms provide primary literature resources  (e.g.  publications), such as Google Scholar\footnote{\url{https://scholar.google.com/}} and Semantic Scholar\footnote{\url{https://www.semanticscholar.org/}}. As an alternative to these advanced materials, other educational platforms such as MOOC.org \footnote{\url{https://www.mooc.org/}} deliver free online courses. Also, unstructured searching on the internet is a popular method to discover other useful resources, such as blog posts, GitHub projects, tutorials, lecture slides and textbooks. Rather than diving into the technical details, these secondary literature resources provide a broad overview of the given domain, which is more valuable for beginners. Still, sifting through this material can be challenging and time-consuming, even if the learner is simply looking for a general and reliable introduction into a new subject area.

\begin{figure*}[th!]
\centering
\includegraphics[width=16cm]{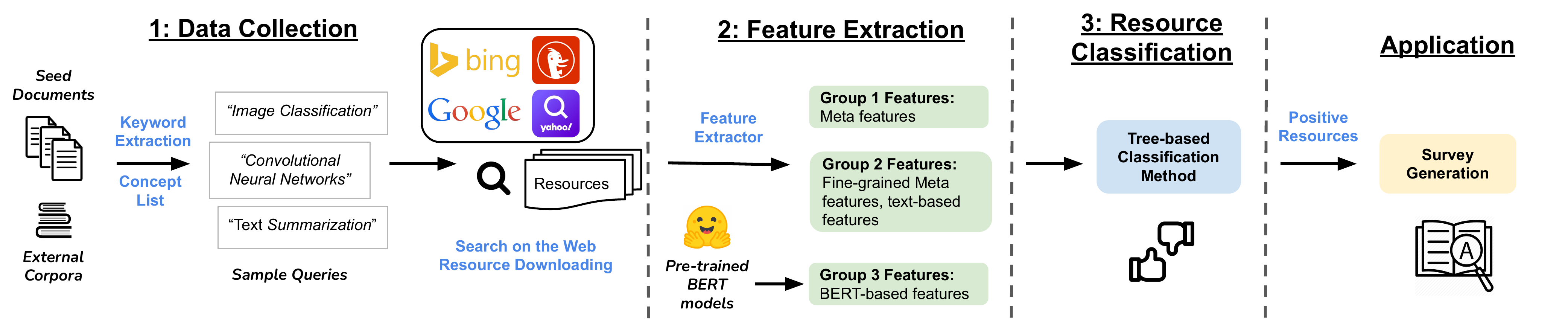}
\caption{Pipeline Overview. The pipeline contains three steps: query generation, feature extraction, and classification \& evaluation. We also show an application in this figure.}
\label{fig:pipeline}
\vspace{-3mm}
\end{figure*}

% related work
Publicly accessible data repositories that focus on gathering a fixed number of educational resources exist, such as scientific papers \cite{Tang:08KDD,Tang:10TKDD}, online platforms AMiner \cite{sinha2015overview} and Semantic Scholar. Some archives also compile secondary literature materials. TutorialBank \cite{fabbri2018tutorialbank} is a manually collected corpus with over 6,300 NLP resources, as well as related fields in Artificial Intelligence (AI), Machine Learning (ML) and so on. LectureBank \cite{li2019whatsi} is also a manually-collected corpus and contains 1,717 lecture slides. MOOCCube \cite{yu2020mooccube} is a large-scale data repository containing 700 MOOC (Massive Open Online Courses). However, in their initial synthesis, these existing corpora either heavily relied on manual efforts that restricted in certain domains, or on a large volume of existing courses sourced from a certain platform. Such solutions are not practically extensible into new or evolving domains. Moreover, according to \cite{fabbri2018tutorialbank}, some web data such as blog posts, tutorials and educational web pages are also suitable materials for learners. These rich web data are ignored. This paper aims to ease the need for human annotators by proposing a pipeline that automates resource discovery to similar unseen domains through transfer learning. Besides, such a pipeline deals with multiple resource types to take advantage of web data.

Our contributions can be summarized into three parts. First, we present a self-sustaining pipeline for educational resource discovery in similar but unseen subject areas or domains. We apply transfer learning with a novel pre-training information retrieval (IR) model, achieving competitive performances. We show that this pipeline achieves 0.94 and 0.82 F1 scores for two arbitrary target domains on discovering high-quality resources. Second, we demonstrate an application that leverage resources discovered by our pipeline, survey generation for the leading paragraph. This is the first study that considers rich web resources for survey generation, to the best of our knowledge. Lastly, we release the core source code of the pipeline and the training and testing datasets, comprised of 39,728 manually labelled web resources and 659 search queries\footnote{\url{https://github.com/yale-lily/Educational-Resource-Discovery}}.

\section{Educational Resource Discovery Pipeline}

We propose the Educational Resource Discovery (ERD) pipeline that aims at automatically recognizing high-quality educational resources. We model this problem as a resource classification task. Given a resource $r$, where $r$ can be any source type such as web page, PDF file and PPTX file, and we can obtain a list of features by feature engineering; based on these features, $r$ is classified as positive if it is a high-quality resource, otherwise negative. We introduce the definition of \textit{high-quality resource} in detail later. We illustrate the ERD pipeline in Figure \ref{fig:pipeline}. It consists of data collection, feature extraction and resource classification.

\subsection{Data Collection} 
\subsubsection{Queries for search}
\label{sec:query}

In this step, we need to conduct a list of meaningful and fine-grain search queries to start. These search queries will then be applied to online search engines for web resources. 
Queries can be borrowed from external corpora or extracted from existing seed documents (e.g., textbooks). We focus on three domains: NLP (natural language processing), CV (computer vision) and STATS (statistics). We utilize external topic lists provided by LectureBankCD \cite{li2021unsupervised}, in which there are 322 NLP-based and 201 CV-based topics from crowdsourcing. For STATS, we extract a list of fine-grained terms from several seed documents, including several textbooks. These terms contain frequent keywords and phrases that are extracted by TextRank \cite{mihalcea2004textrank}, a statistical method to keyword ranking. In total, we end up with 322, 201 and 137 queries for NLP, CV and STATS domains.

To craft our search engine queries, we leverage advanced search conditions: \textit{filetype} and \textit{site} (website). Specifically, we consider three file types: PDF, PPTX/PPT, and HTML. Moreover, according to the TutorialBank corpus \cite{fabbri2018tutorialbank}, resources clustered by the components of their URL possess highly correlated educational content. Thus, we prioritize restricting our queries to websites that consistently provide high-quality resources. We select the top sites from the manually-created TutorialBank corpus and incorporate them into our search queries, as exemplified in \ref{tab:topsite}. We also include the ``.edu'' top-level domain as a special case for our search queries in order to capture general educational resources. Finally, we combine our query terms with the website and filetype constraints: e.g. ``word embeddings filetype:pdf''.  We also augment the original query by generating a disjunction of its variations: e.g., ``stochastic gradient descent'' becomes ``stochastic gradient descent OR SGD''.  Table~\ref{tab:nlpqueries} displays several sample queries.

\begin{table}[t!]
\small
\centering
\begin{tabular}{ll} \toprule
towardsdatascience.com    & datahacker.rs    \\
medium.com                & hackernoon.com\\
www.analyticsvidhya.com   & skymind.ai\\
www.kdnuggets.com         & maelfabien.github.io\\
machinelearningmastery.com & rubikscode.net \\
paperswithcode.com         & research.googleblog.com \\
% datahacker.rs              \\
% hackernoon.com             \\
% skymind.ai                 \\
% maelfabien.github.io       \\
% rubikscode.net             \\
% research.googleblog.com    \\  
\bottomrule
\end{tabular}
\caption{Top sites found in the TutorialBank corpus.}
% .}
\label{tab:topsite}
\vspace{-3mm}
\end{table}

Once the queries are generated, we leverage three well-established online search engines: DuckDuckGo (\url{https://duckduckgo.com/}), Yahoo(\url{https://search.yahoo.com/}) and Bing (\url{https://www.bing.com/}) to obtain our candidate resources. The top $N$ URLs (where $N$ is determined from the domain, file type and site type, varying from 20 to 100 to control the total number of resources we want to collect) for a given query are cached after checking their HTTP response status and ensuring that a URL has not already been collected as part of another query. Moving forward, the documents pointed to by all of these URLs were automatically downloaded and parsed for their features. Certain features, such as the number of authors, were collected using heuristics that accounted for most of the variability within the diverse dataset. The ERD Pipeline's parsers use the pdfminer\footnote{\url{https://github.com/pdfminer/}} and grobid\footnote{\url{https://github.com/kermitt2/grobid}} libraries for PDF files, Apache Tika\footnote{\url{https://tika.apache.org/}} for PPTX/PPT and beautifulsoup\footnote{\url{https://crummy.com/software/BeautifulSoup/}} for HTML.

\subsubsection{Annotation}

After collecting all resources, the next step is to assign a binary label to each resource based on its quality. 
Our annotators consist of 7 graduate and senior college students with a solid background in NLP, CV, and STATS. A resource is annotated as positive if it is a high-quality one. Guidelines for a positive (high-quality) resource are:

\begin{itemize}
    \setlength\itemsep{-0.1em}
    \item \emph{Informative and relevant}: introducing basic knowledge about a specific topic. For example, tutorials, introductions, explanations, guides.
    \item \emph{Papers and lecture slides}: papers and lecture notes about a topic in the correct domain.
    \item \emph{Other secondary literature articles}: i.e., blog posts with informative descriptions, definitions and code blocks.
\end{itemize}

The annotation criteria for a negative (poor-quality) resource are:
\begin{itemize}
    \setlength\itemsep{-0.3em}
    \item \emph{Not informative}: dataset/software/tool download page without introductory descriptions, such as a paper abstract page (not the paper content), a download page with links.
    \item \emph{Irrelevant}: not showing correct content, broken URLs, URLs with not enough or no text (video or image only).
    \item \emph{No knowledge included}: such as a course landing page, a person’s personal website page.
    \item \emph{A list of resources/datasets}: containing only links to other pages.
\end{itemize}

Finally, to measure the inter-coder agreement of the labels, we randomly picked 100 resources and asked each annotator to provide labels independently. Krippendorff's alpha \cite{krippendorff2011computing} on this sample evaluated to 0.8344, indicating a high degree of consistency amongst all annotators.

We detail statistics about our collected dataset in Table \ref{tab:nlpqueries}, providing the total counts by filetype and domain. From the three domains, we collected 39,728 valid resources using 659 distinct queries and achieved a total positive rate of 69.05\%.

\begin{table}[t]
\centering
\footnotesize{
\begin{tabular}{l}
 \toprule
 \textbf{NLP Sample Queries}  \\
``markov decision processes" site:.edu filetype:.pdf\\
``sentiment analysis" site:.edu filetype:.pptx\\
``unlexicalized parsing" site:kdnuggets.com filetype:.html\\
``semantic parsing" site:.edu filetype:.pdf\\
``information retrieval" site:.edu filetype:.pptx \\
``monte carlo methods"  site:rubikscode.net filetype:.html\\
``natural language processing intro" site:.edu filetype:.pdf\\
``sequence to sequence" site:.edu filetype:.pptx \\
``naive bayes" site:paperswithcode.com filetype:.html \\
``latent dirichlet allocation" site:.edu filetype:.pdf\\
\midrule
 \textbf{CV Sample Queries}  \\ 
``epipolar geometry" site:.edu filetype:.pptx\\
``particle filters" site:hackernoon.com filetype:.html\\
``image registration" site:.edu filetype:.pdf\\
``reflectance model" site:.edu filetype:.pptx\\
``shading analysis" site:skymind.ai filetype:.html\\
``imaging geometry and physics" site:.edu filetype:.pdf\\
``texture classification" site:.edu filetype:.pptx\\
``gibbs sampling" site:kdnuggets.com filetype:.html\\
``image thresholding" site:.edu filetype:.pdf\\
``region adjacency graphs" site:.edu filetype:.pptx
\\ \midrule
\textbf{STATS Sample Queries}  \\ 
``linear regression" site:rubikscode.net filetype:.html\\
``hypothesis testing" site:.edu filetype:.pdf \\
``heteroscedasticity" site:.edu filetype:.pptx\\
``random event" site:paperswithcode.com filetype:.html\\
``maximum liklihood" site:.edu filetype:.pdf \\
``granger causality" site:.edu filetype:.pptx\\
``probability" site:hackernoon.com filetype:.html\\
``random sampling" site:.edu filetype:.pdf\\
``correlation coefficient" site:.edu filetype:.pptx\\
``chi-squared statistic" site:skymind.ai filetype:.html\\
\bottomrule
\end{tabular}
}
\caption{Sample queries in the three domains.}
\label{tab:nlpqueries}
\end{table}

\begin{table}[]
\centering
\small
\begin{tabular}{@{}l|rrr|r@{}}
\toprule
        & NLP   & CV    & STATS & \textbf{Total}   \\ \midrule
Query Num & 322   & 200   & 137   & 659   \\ \midrule[0.5pt]
PPTX    & 1,216  & 733   & 1,463  & 3,412  \\
PDF     & 4,961  & 3,782  & 1,449  & 10,192 \\
HTML    & 9,368  & 9,302  & 7,454  & 26,124 \\ 
\textbf{Total}   & 15,545 & 13,817 & 10,366 & \textbf{39,728} \\  \midrule[0.5pt]
Pos.Num & 9,589 &11,101 & 6,742 & 27,432 \\
Pos.Rate & 0.6169 & 0.8034 & 0.6501 &\textbf{0.6905} \\
\bottomrule
\end{tabular}
\caption{Dataset statistics by domain and file type. \textit{Pos.Num} is the number of positive resources. \textit{Pos.Rate} is the fraction of resources that were labeled as positive.}
\label{tab:data}
\end{table}

In Table \ref{tab:data2}, we show token-level and sentence-level statistics on the extracted free text of the collected data.

\begin{table}[h!]
\centering
\small
\begin{tabular}{@{}lrrr@{}}
\toprule
        & NLP   & CV    & STATS  \\ \midrule
\multicolumn{3}{l}{\textit{Token Number/per sentence}} & \\
Mean   & 18.28 & 26.37  & 23.28 \\
Median & 12    & 19     & 18    \\
Max    & 2,302  & 458,363 & 20,066 \\
\midrule[0.3pt]
\multicolumn{3}{l}{\textit{Sentence Number}} & \\
Mean   & 161.60 & 122.49 & 107.32 \\
Median & 55     & 46     & 52     \\
Max    & 5,929   & 21,301  & 52,793  \\
\bottomrule
\end{tabular}
\caption{Free text statistics by domain.}
\label{tab:data2}
\end{table}

% \begin{table}[t]
% \centering
% \small
% \begin{tabular}{c}
%  \toprule
%  \textbf{Sample queries in CV} \\ \midrule
% Image Representation \\
% Image Segmentation \\
% Feature Extraction \\
% Texture Classification \\
% Gibbs Sampling \\
% Image Thresholding \\ 
% Pattern Recognition \\
% Specular Surfaces \\ \bottomrule
% \end{tabular}
% \caption{Sample queries used in CV domain.}
% \label{tab:cvqueries}
% \end{table}

\begin{figure}[t]
\centering
\includegraphics[width=7.5cm]{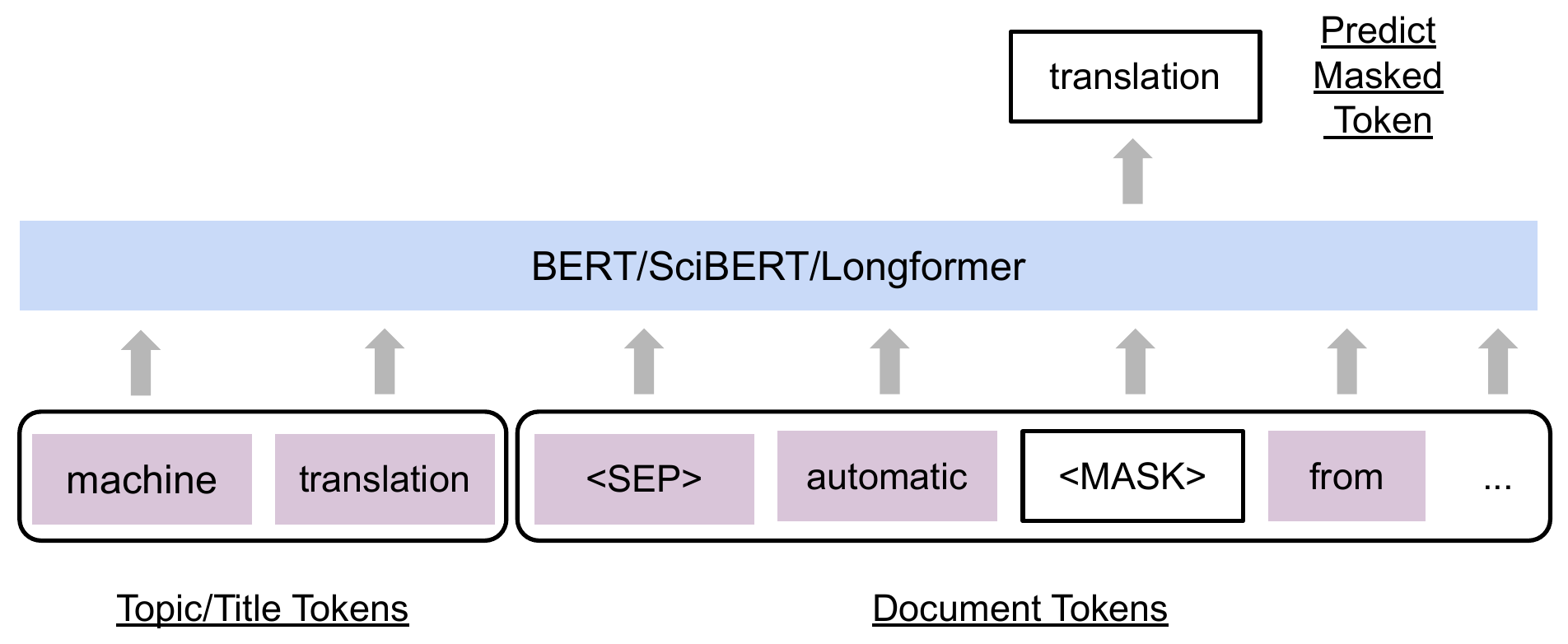}
\caption{QD-MLM pretraining: the topic term \textit{machine translation} is paired with the document tokens. }
\label{fig:qdbert}
\vspace{-3mm}
\end{figure}

\subsection{Feature Extraction}

To train a classifier to identify high-quality educational resources, we first focus on feature engineering. Specifically, we investigate three groups of classification features and summarize them in Table \ref{tab:features}.

\textbf{Group 1 Features} Some of the meta-features of a document that can characterize its quality are embedded in its structure. The features encompassed by Group 1 are high-level and coarse-grained, and focus on aspects such as: the number of headings, equations, outgoing links and authors. Heuristically, some good tutorials may tend to include more equations and paragraphs (longer content).
We list all 8 such features in Table \ref{tab:features}, Group 1. 

\textbf{Group 2 Features} These meta-features describe the fine-grained but statistical details of the resource document. The URL's components, such as the top-level domain name and subdomain name, correlate resources from websites that deliver consistent quality. The other Group 2 features are centered around the characteristics of the free text. For instance,  \textit{NormalizedUniqueVocab} (the size of the vocabulary divided by the total number of words) can estimate the vocabulary's complexity, and \textit{PercentTypos} (the percentage of words that are incorrectly spelled) can approximate reliability. We itemize such features in Table \ref{tab:features}, Group 2.

\begin{table}[t]
\scriptsize
\centering
\begin{tabular}{ll}
\toprule
 \textbf{Feature Name} & \textbf{Explanation} \\ \midrule
 \textit{Group 1} & \vspace{2mm}\\ 
NumAuthor & Number of authors \\ 
NumHeading & Number of headings\\ 
NumFig & Number of figures\\ 
NumEqu & Number of equations \\ 
NumPara & Number of paragraphs \\ 
NumSent & Number of sentences \\ 
NumLink & Number of outgoing links\\ 
BibLen & Bibliography length \\
\midrule
\textit{Group 2} & \vspace{2mm} \\ 
Subdomain	& Subdomain of resource URL  \\
SecondDomain	& Second-level domain of resource URL \\
TopDomain	& Top-level domain of resource URL \\
NumUrlSubdirs	& Number of URL subdirectories \\
NormalizedUniqueVocab	& Number of unique words \\
 & divided by total number of words \\
UniqueVocabMean	& Mean number of occurrences of a word\\
UniqueVocabStdev	& Stdev of number of occurrences of a word \\
WordLenMean	& Mean number of characters per word \\
WordLenStdev	& Stdev of number of characters per word\\
SentenceLenMean	& Mean number of words per sentence \\
SentenceLenStdev	& Stdev of number of words per sentence \\
PercentTypos	& Percentage of words that were misspelled \\
NumGithubLinks	& Number of links to GitHub\\
\midrule
\textit{Group 3} & \vspace{2mm} \\ 
bert &  BERT base model\\
scibert & SciBERT base model \\
longformer & Longformer base model \\
arXiv_bert_QD-MLM & BERT pretrained on arXiv \\
arXiv_scibert_QD-MLM & SciBERT pretrained on arXiv \\
arXiv_longformer_QD-MLM & Longformer pretrained on arXiv \\
TB_bert_QD-MLM & BERT pretrained on TutorialBank \\
TB_scibert_QD-MLM & SciBERT pretrained on TutorialBank \\
TB_longformer_QD-MLM & Longformer pretrained on TutorialBank \\
\bottomrule
\end{tabular}
\caption{Chosen features: we select 3 groups consist of meta features and deep learning-based features.}
\label{tab:features}
\end{table}

\textbf{Group 3 Features} 
The above features are intuitive and statistical features selected by feature engineering. In addition, we introduce contextualized semantic features with pretrained language models as Group 3 features. To achieve this, we choose three base models: BERT \cite{devlin2019bert}, SciBERT \cite{beltagy2019scibert} and Longformer \cite{beltagy2020longformer}. BERT is a language model that was pretrained on Wikipedia documents \footnote{https://huggingface.co/bert-base-uncased}. SciBERT is a BERT-based model trained in the scientific domain, making it suitable for our use case \footnote{https://huggingface.co/allenai/scibert_scivocab_uncased}. Longformer is a BERT-based model that handles longer input sequences \footnote{https://huggingface.co/allenai/longformer-base-4096}. 

We introduce a novel pretraining approach: QD-MLM (Query-Document Masked Language Modeling) to leverage these base models to generate features for our classification task. The model takes a query term and a corresponding document text as an input pair. A query term could be a single word, phrase or a paper title, indicating the \textbf{topic} or \textbf{main idea} of the document.  The pretraining approach is a following of the Masked Language Modeling (MLM) method of BERT (randomly masking 15\% tokens and letting the model predict them), as shown in Figure \ref{fig:qdbert}. To conduct QD-MLM pretraining, 
we apply two external educational corpora for pre-training to ensure the data quality: TutorialBank (TB) \footnote{\url{http://aan.how/download/}} and arXiv \footnote{\url{https://www.kaggle.com/Cornell-University/arxiv}}. The latest TutorialBank has 15,584 topic-document pairs; and arXiv has 259,050 title-abstract pairs (computer science papers only). We enumerate all models in Table \ref{tab:features}, Group 3, naming \textit{dataset}\_\textit{modelname}. We end up with 9 features in Group 3.

After pretraining, the model serves as an encoder for any input documents. However, one may notice that this gives dense embeddings, making it hard to combine with explainable Group 1 and 2 features. Thus, we propose an information retrieval-based scoring function to transfer dense embeddings into a score feature. 
This scoring function calculates a score for each resource, showing the relevancy of the resource to all the searching queries. Relevancy is one of the most important indicators that the resource is annotated as positive. The score is higher if it is more relevant to the queries.  
% The position of a resource in the search results represents its relevance to the learner. Therefore, it plays a central role in the pipeline's ability to support the user's learning and earns its own special feature in our model. In contrast to the all-encompassing online search engines that we extracted the features from, our system solely focuses on educational goals. Namely, we are not concerned with generic content quality, freshness, page speed, brand power and etc, which are major factors in other search engine use cases.
In Section \ref{sec:query}, we apply a list of queries ($q\in Q$) to download resources, we now compute a cosine-similarity based ranking score $score_{r}$ for resource $r$: 
\vspace{-2mm}
\begin{equation}
    \nonumber
    score_{r} = \sum_{q\in Q} cosine\left( V_q, V_r \right)
\end{equation}

where $V_q$ and $V_r$ are QD-MLM encoded embeddings for the query term and resource respectively.

\begin{table*}[t!]
\centering
\begin{tabular}{lccclccc} \toprule 
                              & \multicolumn{3}{c}{\textbf{NLP}$\rightarrow$\textbf{CV}}              &  & \multicolumn{3}{c}{\textbf{NLP}$\rightarrow$\textbf{STATS}}               \\ \cline{2-4} \cline{6-8}
\textbf{Features}             & \textbf{F1}  & \textbf{Precision} & \textbf{Recall} && \textbf{F1}        & \textbf{Precision} & \textbf{Recall} \\ \midrule
Group 1                    & 0.7238  & 0.5802        & 0.9617       && 0.6508  & 0.5405        & 0.8177       \\
Group 1 + 2        & 0.8579  & 0.7772        & 0.9571       && 0.7990  & 0.8141        & 0.7845       \\
% Group 3     &0.7764	&	0.7522&	0.8497&&	0.7923 &	0.7903&	0.7944 \\
Group 1 + 2 + 3 & \textbf{0.9402}  & 0.9849        & 0.8994       && \textbf{0.8225} & 0.9965        & 0.7002  \\
\bottomrule
\end{tabular}
\caption{Classification Results in two target domains: CV and STATS. 
% For Group 3, we report the best model: CV (\texttt{scibert}), STATS (\texttt{TB_scibert}). 
}
\label{tab:res}
\end{table*}

\subsection{Resource Classification}
We treat NLP as the source domain, and CV and STATS as two target domains, given the fact that there are more resources in NLP. So we study two experimental settings: NLP$\to$CV and NLP$\to$STATS.

Since there are various feature types, we conduct prepossessing before applying the classifiers. Numerical values are binned into groups, and categorical features are converted into integer codes. We evaluate four traditional classifiers: Random Forest (RF), Decision Tree (DT), Support Vector Machine (SVM) and Logistic Regression (LR). We find that RF performs the best and has a slight edge over DT, but SVM and LR significantly lag behind. Thus, we report the Random Forest's performance, summarized in Table~\ref{tab:res}. Specifically, we include precision, recall and F1 scores on different feature groups: Group 1, Group 1+2, and Group 1+2+3. It shows that with more features included, better performance can be obtained. When we include all feature groups, we can achieve the best F1 score in both target domains. 
In general, performance on the CV domain is better than on STATS. This is expected given that the corpus distance between NLP and CV is smaller than between NLP and STATS. We give detailed data analysis in the next section.

% The first two rows capture DeText and SBERT in isolation. Concretely, we evaluate four traditional classifiers: Random Forest (RF), Decision Tree (DT), Support Vector Machines and Logistic Regression. We report the best F1 Score for each of the two scenarios. Overall, DeText has a better accuracy and F1 Score in the NLP domain but underperforms in the CV domain. This can be attributed to DeText being fine-tunable during training, while the SBERT layers are frozen. In other words, semantic features are not insightful for when the model transfers from the known NLP domain into the unknown CV domain. Subsequently, we compare the meta features and ranking scores (Meta and Ranking) with the same set of classifiers. Note that the ranking score is simply a floating point value so we combine it with the other Meta Features. Our results show that Random Forest has a slightly higher F1 Score than Decision Tree in both of the domains. The last row shows the classification results with combining all of the features. Decision Tree has the better performance for the NLP domain, while Random Forest for the CV domain. 

\begin{figure}[t]
\centering
  \includegraphics[width=7.2cm]{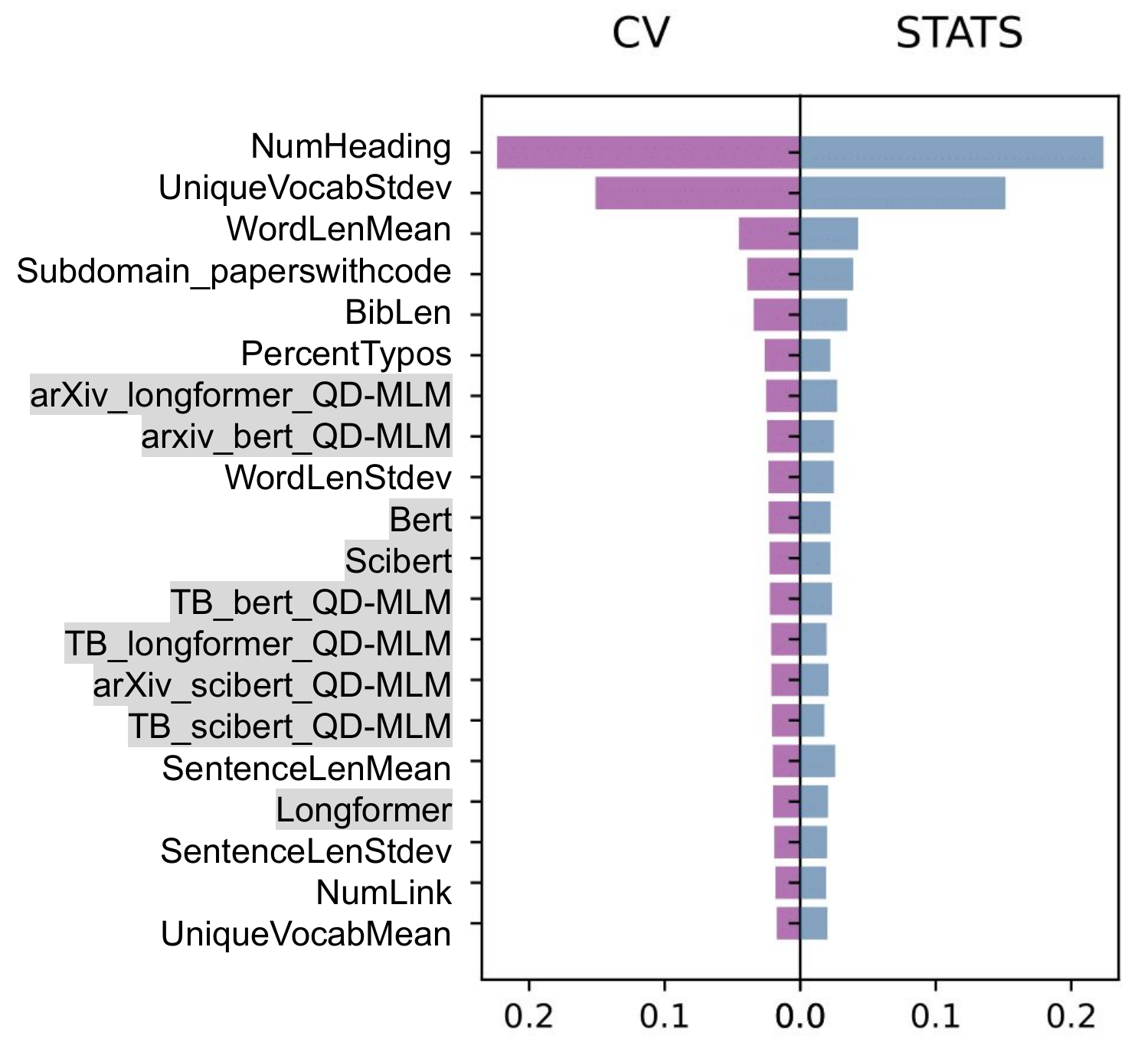}
  \caption{Top 20 features on two target domains: shaded features are from Group 3. }
\label{fig:feat_importance}
\vspace{-3mm}
\end{figure}

% two images
% \begin{figure}[t]
% \centering
% \begin{subfigure}{.5\textwidth}
%   \centering
%   \includegraphics[width=8cm]{side_features0.pdf}
%   \caption{Group 1 + 2:Subdomain_mlm (Subdomain_machinelearningmastery),Subdomain_tds( (Subdomain_towardsdatascience)}
%   \label{fig:featuregroup12}
% \end{subfigure}

% \begin{subfigure}{.5\textwidth}
%   \centering
%   \includegraphics[width=8cm]{side_features.pdf}
%   \caption{All features.}
%   \label{fig:featuresall}
% \end{subfigure}
% \caption{Top 20 features on two domains. }
% \label{fig:feat_importance}
% \end{figure}

% \begin{figure*}[t]
% \centering
% \includegraphics[width=13cm]{cv.png}
% \caption{Top 20 features ordered by importance scores}
% \label{fig:feat}
% \end{figure*}

\section{Data Analysis}

To better understand the collected data and our classifier's performance, we conduct a study on the feature analysis and domain differences. Finally, we also compare similar existing datasets.

\textbf{Feature Importance Score} We take the best-performed model of NLP$\to$CV domain (Group 1+2+3), and take the Gini Index calculated by Decision Trees as the feature importance score. 
Overall, we extract 8746 features in CV and 8525 features of STATS after binning numerical values and encoding categorical features. 
In Figure \ref{fig:feat_importance}, we list the top 20 features of CV and STATS. Some Group 1+2 features rank in the top 5, since they are main indicators that the resource is informative (i.e., more heading numbers, longer contents). 
Some heuristic features such as sentence, word and bibliography lengths rank in the top 20. This observation validates our assumption that such features may indicate more informative content and tend to be a positive resource. 
Additionally, Group 3 features (marked with shaded color) also play an important role. In fact, all Group 3 features rank top 20, suggesting that our scoring function that applies QD-MLM semantic features into the pipeline is very helpful when doing classification for resource discovery.

\begin{table*}[t]
\centering
% \scriptsize
\small
\begin{tabularx}{0.98\textwidth}{lX} \toprule
\textbf{Domain}          & \multicolumn{1}{c}{\textbf{Top 10 Sites}}    \\ \midrule
\textbf{NLP} &\colorbox{clr_cv_font}{www.cs.cmu.edu},
\colorbox{clr_both_font}{web.stanford.edu}, 
\colorbox{clr_cv_font}{www.cs.toronto.edu},
\colorbox{clr_both_font}{www.paperswithcode.com},
\colorbox{clr_both_font}{maelfabien.github.io}, 
\colorbox{clr_cv_font}{www.academia.edu},
\colorbox{clr_cv_font}{courses.cs.washington.edu},
nlp.stanford.edu,
\colorbox{clr_stats_font}{ocw.mit.edu},
\colorbox{clr_cv_font}{www.cs.cornell.edu}\\ \midrule[0.3pt]

\textbf{CV} & www.kdnuggets.com, \colorbox{clr_cv_font}{maelfabien.github.io},\colorbox{clr_cv_font}{www.paperswithcode.com},\colorbox{clr_cv_font}{www.academia.edu}, \colorbox{clr_cv_font}{www.cs.toronto.edu}, \colorbox{clr_cv_font}{www.cs.cmu.edu},\colorbox{clr_cv_font}{web.stanford.edu}, courses.cs.washington.edu, cseweb.ucsd.edu, \colorbox{clr_cv_font}{www.cs.cornell.edu}        \\ \midrule[0.3pt]

\textbf{STATS} & www.kdnuggets.com, \colorbox{clr_stats_font}{maelfabien.github.io}, \colorbox{clr_stats_font}{www.paperswithcode.com},\colorbox{clr_stats_font}{web.stanford.edu},\colorbox{clr_stats_font}{ocw.mit.edu}, online.stat.psu.edu, www.hackernoon.com, www.sjsu.edu, research.googleblog.com,www.cpp.edu     \\
    \bottomrule
\end{tabularx}
\caption{Comparison of the top 10 sites. \textcolor{clr_nlp}{\textbf{Gray}} means overlapped in both CV and STATS domain; \textcolor{clr_cv}{\textbf{Purple}} means overlapping between NLP and CV; \textcolor{clr_stats}{\textbf{Blue}} means overlapping between NLP and STATS.  }
\label{tab:domain}
\end{table*}

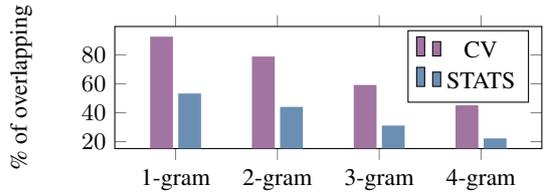
\begin{figure}[t]
\centering
\small
\begin{tikzpicture}
\begin{axis}[
    ybar,
    enlarge x limits={abs=0.8cm},
    bar width=0.3cm,
    legend pos = north east, 
    % legend style={at={(0.5,-0.15)},
    %   anchor=north,legend columns=-1},
    ylabel={\% of overlapping},
    symbolic x coords={1-gram,2-gram,3-gram,4-gram},
    xtick=data,
    % nodes near coords,
    % nodes near coords align={vertical},
      height = 3.2cm,
  width=0.45\textwidth,
    ]
\addplot[fill=clr_cv,draw=none] coordinates {(1-gram, 92.67) (2-gram, 78.92) (3-gram, 59.24) (4-gram,45.17)};
\addplot[fill=clr_stats,draw=none] coordinates {(1-gram, 53.32) (2-gram, 44.09) (3-gram, 31.17) (4-gram, 22.30)};
\addlegendentry{CV}
\addlegendentry{STATS}
\end{axis}
\end{tikzpicture}
\vspace{-2mm}
\caption{Percentage of overlapping n-grams.}
\label{fig:ngrams}
\vspace{-4mm}
\end{figure}

\textbf{Domain Differences} 
As observed before, our transfer learning pipeline performs better on CV than on STATS, and this can be attributed to domain differences relative to the source domain NLP. In Figure \ref{fig:ngrams}, we plot the percentage of overlapping n-grams of both \{NLP, CV\} and \{NLP, STATS\} domain pairs. This indicates that NLP and CV have a larger overlap than \{NLP, STATS\} with respect to all of the n-grams ($n \in \{1,2,3,4\}$). For example, NLP and CV share nearly 80\% on 2-gram, while NLP and STATS share only about 50\%. From this perspective, we uphold that the classifiers trained on semantic features based on BERT models are valuable for bridging more distant domains with transfer learning.

To further contrast our findings, we enumerate the top 10 URLs in Table \ref{tab:domain}. Although the websites are ranked in different orders, there are still common URLs across the domains (highlighted in the table). Once again, CV shares more overlap with NLP in comparison to STATS. 
Along with the feature importance score, this cross-domain consistency further illustrates that the URL meta-features will benefit our model's out-of-domain classification. 
% We leave further investigation and validation of this feature for our future work. 

\textbf{Comparison with Similar Datasets}
We compare a number of existing NLP-related educational datasets in Table \ref{tab:comp_other}, emphasizing the resource type, human effort for annotations, and corpus scale. Note that in this table, we only concentrate on human annotation efforts for free-text resources. This is because these resources are the primary goal of the ERD Pipeline, as opposed to other tasks (e.g. learning concept relations, concept mining).
We can see that MOOCcube \cite{yu2020mooccube} has a massive quantities of a single resource type (papers). They obtained the metadata from a third-party platform, AMiner, without a full round of human annotations. TutorialBank \cite{fabbri2018tutorialbank} has a larger number of resources than LectureBank \cite{li2019whatsi}, and it consists of diverse resource types. Our pipeline is very similar to TutorialBank in terms of resource type, but ours extends to more resources and subject areas, enabling us to research transfer learning across domains.

\begin{table*}[t]
\centering
% \footnotesize
\small
\begin{tabular}{lcccr} \toprule
    \textbf{Name}     & \textbf{Resource Type (with texts)}     & \textbf{Domain Number}                   & \textbf{Annotation} & \textbf{Size} \\ \midrule
TutorialBank  & \textbf{Lecture sides, papers, blog posts} & NLP only & \textbf{Manually}                     &  6,300    \\
LectureBank  & Lecture sides only        &   NLP only       & \textbf{Manually}                     &  1,717    \\
MOOCcube    & Papers only                  &    \textbf{Multiple}       & Scrape from third-party  &   679,790   \\
ERD (ours)         & \textbf{Lecture sides, papers, blog posts}   &    \textbf{Multiple}    & 
\textbf{Manually}           &   39,728  \\ \bottomrule
\end{tabular}
\caption{Comparison with similar datasets.}
\label{tab:comp_other}
% \vspace{-3mm}
\end{table*}

\section{Application: Leading Paragraph Generation for Surveys}

This section demonstrates an interesting application that applies the resources discovered using our ERD Pipeline: leading paragraph generation for surveys using web data.

Novel concepts are being introduced and evolving at a rate that creates high-quality surveys for web resources, such as Wikipedia pages, challenging. Such existing surveys like Wikipedia still needs human efforts on collecting relevant resources and writing accurate content on a given topic. 
% Researchers have been investigating automatic ways to generate surveys using machine learning and deep learning methods. 
Survey generation is a way to generate concise introductory content for a query topic \cite{DBLP:journals/csl/ZhaoYLZHCNLG21}. 
While most of the existing work focuses on utilizing scientific papers or Wikipedia to achieve this \cite{liu2018generating}, little has been done for using the data on the web. 
Since our ERD pipeline discovers sufficient web data, we propose a two-stage approach for generating the lead paragraph that applies these web data. Notably, this is the first research on abstractive survey generation using web data.

\begin{table}[b]
\begin{subtable}{0.5\textwidth}
\centering
\small
\resizebox{\columnwidth}{!}{\begin{tabular}{c c c c c} \Xhline{2\arrayrulewidth}
             \specialcell{Methods} & \textbf{L=5} & \textbf{L=10} & \textbf{L=20} & \textbf{L=40} \\ \Xhline{\arrayrulewidth}
            %  TF-IDF & 24.86 & 32.43 & 40.87 &  49.49  \\  
              LSTM-Rank & 39.38 &  46.74 &  53.84 & 60.42  \\  
            %   WikiCite & \textbf{65.27} & 69.77 &  73.54 & 76.51  \\  
              Semantic Search & 34.87 & 48.60 & 61.87 & 74.54  \\ 
              RoBERTa-Rank & \textbf{64.12} & \textbf{72.49} & \textbf{79.17} & \textbf{84.28}  \\ \Xhline{2\arrayrulewidth}
\end{tabular}}
\caption{ROUGE-L \cite{lin-2004-rouge} Recall scores for WikiSum content selection, varying the number of paragraphs returned.}
\label{tab:content_selection}
\end{subtable}

\bigskip
\begin{subtable}{0.5\textwidth}
\centering
\small
\begin{tabular}{c c c c } \Xhline{2\arrayrulewidth}
             \specialcell{Methods} & \textbf{R-1} & \textbf{R-2} & \textbf{R-L} \\ \Xhline{\arrayrulewidth}
             HierSumm \tiny{\cite{liu2019hierarchical}} & 41.53 & 26.52 & 35.76 \\  
              BART \tiny{\cite{lewis2019bart}} & \textbf{46.61} &  \textbf{26.82} &  \textbf{43.25} \\   \Xhline{2\arrayrulewidth}
\end{tabular}
\caption{ROUGE scores for intro generation.}
\label{tab:text_generation}
\end{subtable}

\caption{Two-stage method evaluation using WikiSum.} \label{tab:three_tables}
\end{table}

\subsection{Two-stage method}
\begin{figure}[t]
\centering
\includegraphics[width=0.45\textwidth]{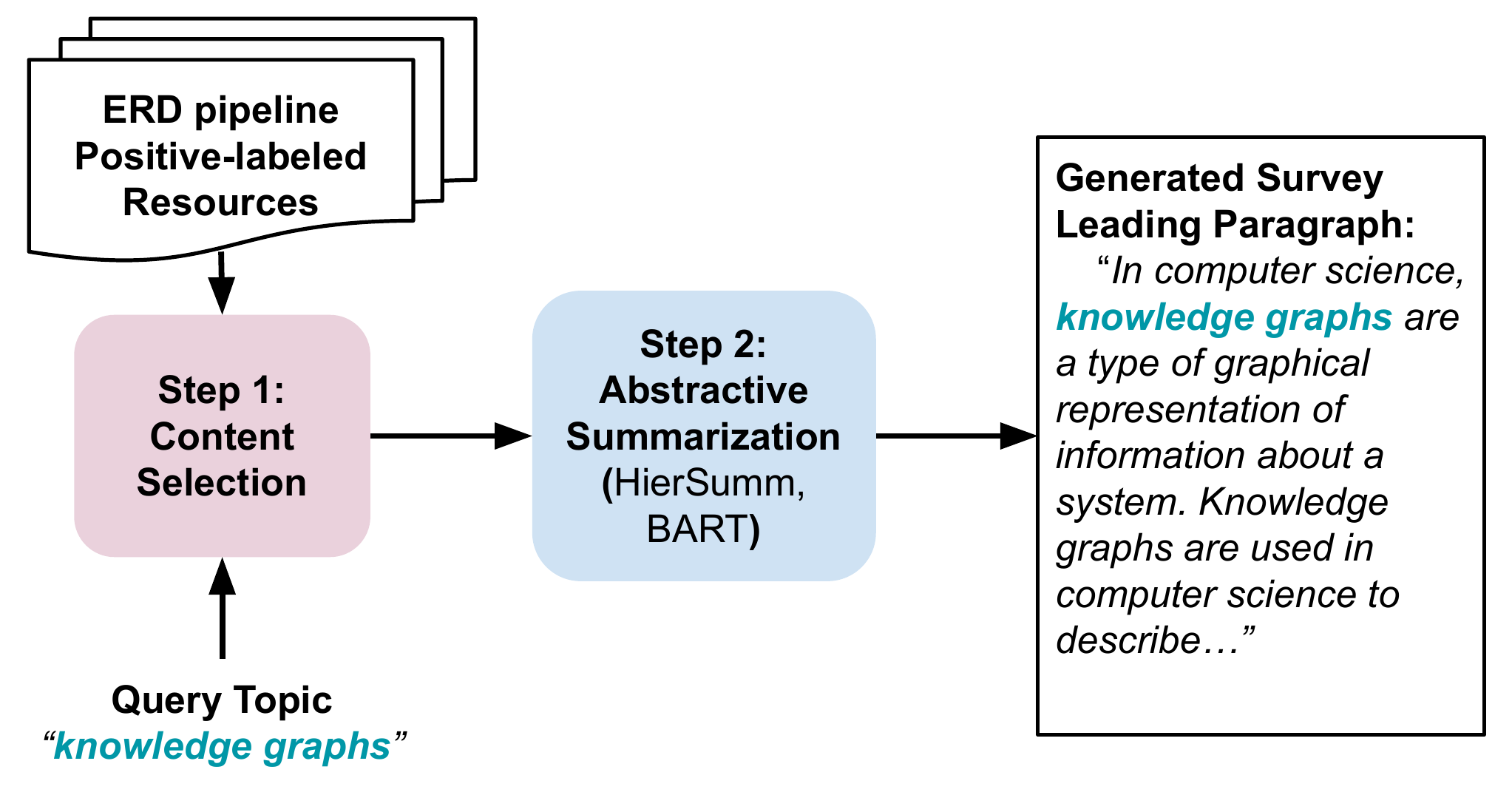}
\caption{Two-stage Method for Leading Paragraph Generation. The query topic is \textit{knowledge graph} as an example. }
\label{fig:surgen}
\end{figure}

We propose a two-stage method illustrates in Figure \ref{fig:surgen}. Given a query topic and high-quality web resources selected by the ERD pipeline, we wish to generate the leading introductory paragraph for the query topic. 
This approach consists of content selection (step 1) and abstractive summarization (step 2). Content selection is the process of selecting the most relevant sentences according to the given query.  Abstractive summarization generates the accurate lead paragraph from the selected sentences.

\textbf{Content Selection}
ERD is supposed to identify massive resources with broad coverage of the topics, so the first step is to select related content with the query topic. 

While there is no suitable pretrained data for this survey generation task, we utilize the WikiSum dataset \cite{liu2018generating}. WikiSum contains 1.5 million Wikipedia pages, their references and their associated Google Search results. WikiSum includes many well-established topics and comprehensive reference documents, making it suitable for survey generation. We first evaluate content selection models using WikiSum. 
We experiment with three approaches in this step. \newcite{liu2019hierarchical} undertake query-based content selection as a regression problem of predicting the ROUGE-2 recall of a given paragraph-topic pair (\texttt{LSTM-Rank}). 
\newcite{reimers2019sentence} fine-tune BERT \cite{devlin2019bert} and RoBERTa \cite{liu2019roberta} to produce fixed-length vectors which can be compared using cosine similarity. We embed the topic of each Wikipedia page and candidate paragraph using this method, and select the paragraphs with the closest vectors to the title (\texttt{Semantic Search}).
Additionally, we train RoBERTa in a similar manner as \cite{liu2019hierarchical}. Then, we compare the query topic and paragraphs as sentence pairs and use the resulting relevance scores for the paragraph ranking (\texttt{RoBERTa-Rank}).
As shown in Table \ref{tab:content_selection}, RoBERTa-Rank is the highest-scoring content selector, so we employ it for the abstractive summarization's input.

\begin{table}[b]
\centering
\small
\resizebox{\columnwidth}{!}{\begin{tabular}{cccc}
\toprule
    \textbf{Evaluation} & NLP & CV & STATS \\ \midrule
    Avg. Readability & 3.45 & 2.90 & 2.75 \\
    Avg. Relevance & 2.80 & 1.85 & 1.50 \\
    Avg. Non-redundancy & 2.45 & 2.15 & 1.60 \\ \bottomrule
\end{tabular}}
\caption{Human Evaluation}
\label{tab:final_analysis}
\end{table}

\textbf{Abstractive Summarization}
This step is to generate summarization from the content selected previously. As a sequence-to-sequence task, there are many existing pretrained models. We experiment with \texttt{BART} \cite{lewis2019bart}, a pretrained model for text generation, as well as \texttt{HierSumm}, a hierarchical model from \newcite{liu2019hierarchical}. We show the summarization results on the WikiSum data in Table \ref{tab:text_generation}, and observe that BART achieves a better performance.

% Besides, by using the RoBERTa content selection component, we are able to select paragraphs up to 1024 total tokens. 

\subsection{Human Evaluation and Case Studies}

So far, we have shown that applying RoBERTa-Rank and BART as a two-stage method gives promising results evaluated on the WikiSum dataset. We connect our pipeline with this method to generate the leading paragraph. We choose ten queries randomly as survey topics in each domain, for example, \lq\lq sentiment analysis\rq \rq in NLP. A complete query topic list can be found in Table \ref{tab:ten_topics}. Since we do not have ground truth, we conduct human evaluation and case studies.

\begin{table}[htb!]
\centering
\footnotesize{
\begin{tabularx}{0.48\textwidth} {X} \toprule
 \textbf{NLP}: adam optimizer, 
lstm model, 
dropout neural networks, 
recursive neural network, 
convolutional neural network, 
automatic summarization, 
sentiment analysis, 
attention mechanism deep learning, 
Pre-trained Language Models NLP, 
knowledge graphs\\
\midrule
 \textbf{CV}: 
transfer learning, 
convolutional neural network, 
image retrieval, 
image classification, 
feature learning, 
seq2seq, 
transformers, 
visual question answering, 
conditional probability, 
k means	\\
\midrule
\textbf{STATS}: 
linear regression, 
hypothesis testing, 
conditional probability, 
multinomial distribution, 
probability density, 
density estimation, 
normal distributions, 
bernoulli distribution, 
standard deviation, 
z-score	 \\
\bottomrule
\end{tabularx}
}
\caption{Topics selected for human evaluation.}
\label{tab:ten_topics}
\end{table}

\begin{table*}[t]
\centering
\footnotesize
\begin{tabularx}{0.95\linewidth} {X} \toprule
%  \textbf{Domain} & \textbf{Survey Concept} & \textbf{Generated Survey} \\ \midrule
 \textbf{NLP: knowledge graphs} \\
 \:\:\:\:In computer science, knowledge graphs are a type of graphical representation of information about a system. Knowledge graphs are used in computer science to \underline{describe the structure of a system}, such as a graph of knowledge about a particular system, ...
 or a list of items that can be used to describe a system 's structure. For example, a knowledge graph may contain information about an object 's shape, its size, its shape relative to other objects, its orientation, and its orientation relative to a point in space, or its position relative to another point in time, or any combination of these things. 
 Knowledge Graphs can also be used as a way of \underline{describing the structure and structure of systems}, such that a system can be described by a set of knowledge graphs that \underline{describe its structure} and its properties.\\
\midrule[0.5pt]
 \textbf{CV: image retrieval} \\
 \:\:\:\:In computer science, a Convolutional neural network is a type of neural network used in image analysis and image synthesis. It is a computer program that uses a neural network to learn from a large number of images, and can be used to predict the appearance of a given image in real-world situations, such as the shape of a surface, or the color of an object in a 3D environment such as an image of a person 's face. It can also be used for image synthesis and image analysis, as well as image processing and image processing for computer vision and image recognition. It has been used in a number of applications such as computer vision, image processing, image recognition, and computer vision for image recognition and image rendering.\\ \midrule[0.5pt]
 \textbf{STATS: normal distributions}\\
 \:\:\:\:In physics, normal distributions are a family of mathematical models that describe the distribution of normal distributions. They are used in the fields of statistics, physics , and computer science, and have been used in a wide variety of applications, including computer vision, image analysis, \underline{computer graphics}, computer vision, \underline{computer graphics processing units} (CPGs), as well as in the field of computer vision. They have also been used to study the visual appearance of real-world surfaces, such as the Phong Reflection Model, the Oren-Nayar model, the Koenderink et al. representation, and the Shading of the Sphere Model, and in the study of light sources and light sources in OpenGL and OpenGL-based graphics renderers.  \\
 \bottomrule
\hline
\end{tabularx}
\caption{Examples of generated leading paragraphs. }
\label{tab:survey_example}
\end{table*}

We evaluate the model outputs on a 1-5 Likert scale based on the following qualities: 
% \emph{Readability}: attains a maximum score of 5 if the output is readable with a high degree of fluency and coherency.
% \emph{Relevancy}: attains a maximum score of 5 if the output is perfectly relevant to the current topic with no hallucinations.
% \emph{Non-redundancy}: attains a maximum score of 5 if the output has no repeating phrases/concepts.
\begin{itemize}
    \setlength\itemsep{-0.1em}
    \item \emph{Readability}: attains a maximum score of 5 if the output is readable with a high degree of fluency and coherency.
    \item \emph{Relevancy}: attains a maximum score of 5 if the output is perfectly relevant to the current topic with no hallucinations.
    \item \emph{Non-redundancy}: attains a maximum score of 5 if the output has no repeating phrases/concepts.
\end{itemize}

We report average scores among 2 human judges of all topics by domain, shown in Table~\ref{tab:final_analysis}. The scores of NLP are the highest for all qualities, and STATS performed most poorly. This discrepancy may be caused by data collection bias, as more NLP resources were included. 
%This is the same trend with the resource discovery experiments. 

We randomly pick one case study from each domain in Table~\ref{tab:survey_example}. We eliminate some parts of the generated content due to limited space. The model is able to generate leading paragraphs in a similar Wikipedia article style by giving a definition of a certain concept, following by descriptions of possible applications. Overall, while these surveys contains some facts, but the quality can still be improved. For instance, the STATS paragraph exhibits some repetition and redundancy (e.g., ``computer graphics'',``computer graphics processing units''). This also happens in the NLP topic: ``describe the structure of a system''). We highlight them in underline. As an initial study, we have demonstrated the opportunities of extending our ERD Pipeline to produce survey paragraphs. In the future, we aim to enhance the generated leading paragraphs and extend the model for generating complete surveys.

\section{Conclusion}

In this paper, we proposed a pipeline for automatic knowledge discovery in novel domains. We applied transfer learning with an MLM pre-training method and achieved competitive classification performances. We demonstrated an application to take advantage of resources discovered by our pipeline. Finally, we released our source code and the collected datasets, including the 39,728 manually labelled web resources and 659 search queries. We make this pipeline an online live educational tool for the public.

% \nocite{*}
\section{Bibliographical References}\label{reference}
%\label{main:ref}

\bibliographystyle{lrec2022-bib}
\bibliography{lrec2022-example}

% \section{Language Resource References}
% \label{lr:ref}
% \bibliographystylelanguageresource{lrec2022-bib}
% \bibliographylanguageresource{languageresource}

\end{document}